\documentclass[10pt]{llncs}
\usepackage{rotating}
\usepackage{pdflscape}
\usepackage{amsmath}
\usepackage{amsfonts}
\usepackage{amssymb}
\usepackage{amstext}
\usepackage{latexsym,stmaryrd}
\usepackage[ruled, linesnumbered,noend]{algorithm2e}

\newtheorem{lemme}{Lemma}
\newtheorem{propriete}{Property}
\newtheorem{preuve}{Proof}
\usepackage{multirow}
\usepackage{colortbl}
\usepackage{xcolor}
\usepackage{inferance}
\usepackage{deduc}

\def\infty{\text{BUG}}

\renewcommand\ruleformat[1]{\ #1}

\begin{document}
\title{Towards Learned Clauses Database Reduction Strategies Based on Dominance Relationship}
\titlerunning{Learned Clauses Database Reduction}
\author{Jerry Lonlac$^{1,2}$  \and  Engelbert Mephu Nguifo$^1$}

\institute{$^1$ Clermont Auvergne University, CNRS, LIMOS, F-63000 Clermont-Ferrand, France \\
$^2$ Clermont Auvergne University, CNRS, GEOLAB, F-63000 Clermont-Ferrand \\
\{jerry.lonlac\_konlac, engelbert.mephu\_nguifo\}@uca.fr}

\maketitle

\begin{abstract}

Clause Learning is one of the most important components of a conflict driven clause learning (CDCL) SAT solver that is effective on industrial instances. Since the number of learned clauses is proved to be exponential in the worse case, it is necessary to identify the most relevant clauses to maintain and delete the irrelevant ones. As reported in the literature, several learned clauses deletion strategies have been proposed. However the diversity in both the number of clauses to be removed at each step of reduction and the results obtained with each strategy creates confusion to determine which criterion is better. Thus, the problem to select which learned clauses are to be removed during the search step remains very challenging. In this paper, we propose a novel approach to identify the most relevant learned clauses without favoring or excluding any of the proposed measures, but by adopting the notion of dominance relationship among those measures. 
Our approach bypasses the problem of the diversity of results and reaches a compromise between the assessments of these measures. Furthermore, the proposed approach also avoids another non-trivial problem which is the amount of clauses to be deleted at each reduction of the learned clause database.

\end{abstract}

\section{Introduction}
The SAT problem, i.e., the problem of checking whether a Boolean formula in conjunctive normal form (CNF) is satisfiable or not, is central to many domains in computer science and artificial intelligence including constraint satisfaction problems (CSP), automated planning, non-monotonic reasoning, VLSI correctness checking, etc. Today, SAT has gained a considerable audience with the advent of a new generation of solvers able to solve large instances encoding real-world problems. 
These solvers, often called \emph{modern SAT solvers} \cite{Moskewicz01,MiniSat03} or CDCL (Conflict Driven Clause Learning) SAT solvers have been shown to be very efficient at solving real-world SAT instances. 
They are built by integrating four major components to the classical (DPLL) procedure \cite{Davis62}: lazy data structures \cite{Moskewicz01}, activity-based variable selection heuristics (VSIDS-like) \cite{Moskewicz01}, restart policies \cite{Gomes1998}, and clause learning \cite{Marques-SilvaS99,Moskewicz01}. Although a nice combination of these components contributes to improve the efficiency of modern SAT solvers \cite{KatebiSS11}, clause learning is known as the most important component \cite{PipatsrisawatD09}.
The global idea of clause learning is that during the unit propagation process, when a current branch of the search tree leads to a conflict, moderns SAT solvers learn a conflict clause that helps unit propagation to discover one of the implications missed at an earlier level. This conflict clause expresses the causes of the conflict and is used to prune the search space.
Clause learning, also known in the literature as Conflict Driven Clause Learning (CDCL), refers now to the most known and used First (UIP) learning scheme, first integrated in the SAT solver Grasp \cite{Marques-Silva96} and efficiently implemented in zChaff \cite{Moskewicz01}. Most of the SAT solvers integrate this strong learning scheme.
Since at each conflict, CDCL solvers learn a new clause that is added to the learned clauses database, and the number of learned clauses is proved to be exponential in the worse case, it is necessary to remove some learned clauses to maintain a database of polynomial size. Therefore, removing too many clauses can make learning inefficient, and keeping too many clauses also can alter the efficiency of unit propagation.

Managing the learned clauses database was the subject of several studies \cite{Moskewicz01,Marques-Silva96,MiniSat03,AudemardS09,AudemardLMS11,GuoJLS14}. These strategies were proposed with the objective to maintain a learned clause database of reasonable size by eliminating clauses deemed irrelevant to the subsequent search. The general principle of these strategies is that, at each conflict, an activity is associated to the learned clauses (static strategy). Such heuristic-based activity aims to weight each clause according to its relevance to the search process. In the case of dynamic strategies, such clauses activities are dynamically updated. The reduction of the learned clauses database consists in eliminating inactive or irrelevant clauses. 
Although all the learned clause deletion strategies proposed in the literature
are shown to be empirically efficient, identifying the most relevant clause to maintain during the search process remains a challenging task.
Our motivation in this work comes from the observation that the use of different relevant-based deletion strategies gives different performances.
Our goal is to take advantage of several relevant learned clauses deletion strategies by seeking a compromise between them through a dominance relationship.

 In this paper, we integrate a user-preference point of view in the SAT process. To this end, we integrate into the SAT process the idea of skyline queries \cite{BorzsonyiKS01}, dominant patterns \cite{SouletRPC11}, undominated association rules \cite{BoukerSYN14} in order to learn clauses  in a threshold-free manner. Such queries have attracted considerable attention due to their importance in multi-criteria decision making.  Given a set of clauses, the skyline set contains the clauses that are not dominated by any other clause.

Skyline processing does not require any threshold selection function, and the formal property of domination satisfied by the skyline clauses gives to the clauses a global interest with semantics easily understood by the user. This skyline notion has been developed for database and data mining applications, however it was unused for SAT purposes. In this paper, we adapt this notion to the learned clauses management process.

The paper is organized as follows. We first present some effective relevant-based learned clauses deletion strategies used in the literature. Then, our learned clauses deletion strategy based on the dominance relationship between different strategies
is presented in section \ref{sec:detectUndominatedClause}.
Finally, before the conclusion, experimental results demonstrating the efficiency of our approach are presented.

\section{On the learned clauses database management strategies}
\label{sec:measure}
In this section, we present some efficient learned clauses relevance measures exploited in the most SAT solvers of the literature.


The most popular CDCL SAT solver \textit{Minisat} \cite{MiniSat03} considers as relevant the clauses the most involved in recent conflict analysis and removes the learned clauses whose involvement in recent conflict analysis is marginal. Another strategy called \textit{LBD} for \textit{Literal Block Distance} was proposed in \cite{AudemardS09}. LBD based measure is also exploited by most of the best state-of-the-art SAT solver (\textit{Glucose, Lingeling \cite{Lingeling12}}) and whose efficiency has been proved empirically. LBD based measure uses the number of different levels involved in a given learned clause to quantify the quality of the learned clauses. Hence, the clauses with smaller LBD are considered as more relevant. 
In \cite{AudemardLMS11}, a new dynamic management policy of the learned clauses database is proposed. It is based on a dynamic freezing and activation principle of the learned clauses. At a given search state, using a relevant selection function based on progress saving (PSM), it activates the most promising learned clauses while freezing irrelevant ones.
In \cite{GuoJLS14}, a new criterion to quantify the relevance of a clause using its backtrack level called BTL for BackTrack Level was proposed. From experiments, the authors observed that the learned clauses with small BTL values are used more often in the unit propagation process than those with higher BTL values. More precisely, the authors observed that the learned clauses with BTL value less than 3 are always used much more than the remaining clauses. Starting from this observation, and motivated by the fact that a learned clause with smaller BTL contains more literals from
the top of the search tree, the authors deduce that relevant clauses are those allowing a higher backtracking in the search tree (having small BTL value). 
More recently, several other learned clauses database strategies were proposed  in \cite{JabbourLSS14,AnsoteguiGLS15}. In \cite{JabbourLSS14}, the authors explore a number of variations of learned clause database reduction strategies, and the performance of the different extensions of \textit{Minisat} solver integrating their strategies is evaluated on the instances of the SAT competitions 2013/2014 and compared against other state-of-the-art SAT solvers (\textit{Glucose, Lingeling}) as well as against default \textit{Minisat}.
From the performances obtained in \cite{JabbourLSS14}, the authors have shown that size-bounded learning strategies proposed more than fifteenth years ago \cite{Marques-Silva96,Bayardo96acomplexity,Bayardo97} is not over and remains a good measure to predict the quality of learned clauses. They show that adding randomization to size bounded learning is a nice way to achieve controlled diversification, allows to favor the short clauses, while maintaining a small fraction of large clauses necessary for deriving resolution proofs on some SAT instances.
This study opens many discussions about the learned clauses database strategies and raises questions about the effectiveness proclaimed by other strategies of the state-of-the-art \cite{MiniSat03,AudemardS09}. 
In \cite{AnsoteguiGLS15}, the authors use the community structure of industrial SAT instances to identify a set of highly useful learned clauses. They show that augmenting a SAT instance with the clauses learned by the solver during its execution does not always mean to make the instance easy. However, the authors show that augmenting the formula with a set of clauses based on the community structure of the formula improves the performance of the solver in many cases.
The different performances obtained by each strategy suggests that the question on how to predict efficiently the "best" learned clauses is still open and deserves further investigation.

On the other hand, it is important to note that the efficiency of most of these state-of-the-art learned clauses  management strategies heavily depends on the cleaning frequency and on the amount of clauses to be deleted each time. Generally, all the CDCL SAT solvers using these strategies exactly delete half of the learned clauses at each learned clauses database reduction step. For example, the CDCL SAT solver \textit{Minisat} \cite{MiniSat03} and \textit{Glucose} \cite{AudemardS09} delete half of the learned clauses at each cleaning.
Therefore, the efficiency of this amount of learned clauses to delete (e.g the half) at each cleaning step of the learned clauses database has not been demonstrated theoretically, but instead experimentally. For our knowledge, there are not many studies in the literature on how to determine the amount of  clauses to be deleted each time. This paper proposes an approach to identify the relevant learned clauses during the resolution process without favoring any of the best reported relevant measures  and which frees itself of the amount of clauses to be removed at each time: the amount of learned clauses to delete corresponds at each time to the number of learned clauses dominated by one particular learned clause of the set of the current learned clauses which is called in the following sections, the reference learned clause.

\section{Detecting undominated learned Clauses} 
\label{sec:detectUndominatedClause}
We present now our learned clauses relevant measure based on dominance relationship. 
We first motivate this approach with a simple example, and then propose an algorithm allowing to identify the relevant clauses with some technical details.
\subsection{Motivating example}
\label{sec:motivating}
Let us consider the following relevant strategies: \textit{LBD} \cite{AudemardS09}, \textit{SIZE} (which consider as relevant the clause of the short size) and the relevant measure use by \textit{minisat} \cite{MiniSat03} that we denote here \textit{CVSIDS}.
Suppose that we have in the learned clauses database, the clauses $c_1$, $c_2$ and $c_3$ with:
\begin{footnotesize}
\begin{itemize}
\item $SIZE(c_1) = 8$, $LBD(c_1) = 3$, $CVSIDS(c_1) =  1e^{100}$; 
\item $SIZE(c_2) = 6$, $LBD(c_2) = 5$, $CVSIDS(c_2) =  1e^{200}$;  
\item $SIZE(c_3) = 5$, $LBD(c_3) = 4$, $CVSIDS(c_3) =  1e^{300}$. 
\end{itemize}
\end{footnotesize}
The question we ask is the following: which one is relevant? 
In \cite{AudemardS09}, the authors consider the clause $c_1$ which has the most smallest LBD measure as the most relevant. In contrast, the authors of \cite{JabbourLSS14} and \cite{Goldberg20071549} prefer the clause $c_3$ while the preference of the authors of \textit{Minisat} \cite{MiniSat03} leads to the clause $c_3$.  Our approach copes with the particular preference at one measure by finding a compromise between the different relevant measures through the dominance relationship. Hence, for the situation described above, only the clause $c_2$ is irrelevant because it is dominated by the clause $c_3$ on the three given measures.

\subsection{Formalization}
\label{sec:formalization}
During the search process, the CDCL SAT solvers learn a set of clauses which are stored in the learned clauses database $\Delta$, $\Delta = \{c_1, c_2, . . ., c_n\}$. At each cleaning step, we evaluate these clauses with respect to a set $\cal{M}$ $= \{m_1, m_2, . . ., m_k\}$ of relevant measures. We denote $m(c)$ the value of the measure $m$ for the clause  $c$, $c \in \Delta$, $m \in \cal{M}$.
Since the evaluation of learned clauses varies from a measure to another one, using several measures could lead to different outputs (relevant clauses with respect to a measure).
For example, if we consider the motivating example, $c_1$ is the best clause with respect to the $LBD$ measure whereas it is not the case according to the evaluation of $SIZE$ measure which favors $c_3$ . This difference of evaluations is confusing for any process of learned clauses selection.
Hence, we can utilize the notion of dominance between learned clauses to address the selection of relevant ones. Before, formulating the dominance relationship between learned clauses, we need to define it at the level of measure values. To do that, we define dominance value as follows:

\begin{definition}[dominance value]
\label{def:domvalue}
Given a learned clauses relevant measure $m$ and two learned clauses $c$ and $c'$, we say that $m(c)$ dominates $m(c')$, denoted by $m(c) \succeq m(c')$, iff $m(c)$ is preferred to $m(c')$. If $m(c) \succeq m(c')$ and $m(c) \neq m(c')$ then we say that $m(c)$ strictly dominates m(c'), denoted $m(c) \succ m(c')$.
\end{definition}

\begin{definition}[dominance clause]
\label{def:dominanceclaue}
Given two learned clauses  $c, c'$, the dominance relationship according to the set of learned clauses relevant measures $\cal{M}$ is defined as follows:
\begin{itemize}
\item $c$ dominates $c'$, denoted $c \succeq c'$,  iff $m(c) \succeq m(c')$, $\forall m \in \cal{M}$.
\item If $c$ dominates $c'$ and $\exists m \in \cal{M}$ such that $m(c) \succ m(c')$, then $c$ stritly dominates $c'$ and we note  $c \succ c'$.
\end{itemize}
\end{definition}

To discover the relevant learned clauses a naive approach consists in comparing each clause with all other ones. However, the number of learned clauses is proved to be exponential which makes pairwise comparisons costly. In the following, we show how to overcome this problem by defining at each cleaning step of learned clauses database, a particular learned clause denoted by $\tau$ that we call here \textit{current reference learned clause} which is an undominated clause of $\Delta$ according to the set of learned clauses relevant measures $\cal{M}$. At each cleaning step, all the learned clauses dominated by $\tau$ will be considered as the irrelevant learned clauses and thus deleted from the learned clauses database.

To define \textit{current reference learned clause}, we need a new relevant measure based on all the learned clauses relevant measures of $\cal{M}$. We call this new measure \textit{Degree of compromise}, in short \textit{DegComp} defines as follows:

\begin{definition}[Degree of compromise]
\label{def:DegComp}
Given a learned clause $c$, the degree of compromise of $c$ with respect to the set of learned clauses relevant measures $\cal{M}$ is defined by $DegComp(c) = \frac{\sum_{i=1}^{n}\widehat{m_i(c)}}{|M|}$, where $\widehat{m_i(c)}$ corresponds to the normalized value of the clause $c$ on the measure $m_i$. 
\end{definition}

In fact, in practice, measures are heterogeneous and defined within different scales.
For example the values of the learned clauses relevant measures in \cite{MiniSat03} are very high, in exponential order while the values of the relevant measures in \cite{AudemardS09} are smallest ones. Hence, in order to avoid that the measures with the higher values make marginal the measures with smallest values in the computation of the comprise degree of a given learned clauses, it is recommended to normalize the measures values. In our case here, we choose to normalize all the measures in the interval $[0,1]$. 
More precisely, each value of measure $m(c)$ of any learned clause $c$ must be normalized into $\widehat{m(c)}$ within $[0,1]$. The normalization of a given measure $m$ is performed
depending on its domain and the statistical distribution of its active domain. We recall that the active domain of a measure $m$ is the set of its possible values.  It is worth mentioning, \textbf{the normalization of a measure does not modify the dominance relationship} between two given values. If we consider the learned clause $c_1$ given in the motivating example in the section \ref{sec:motivating}, with its three values : $DegComp(c_1)= \frac{\widehat{CVSDIS(c_1)} + \widehat{LBD(c_1)} + \widehat{SIZE(c_1)}}{3}$, then, we have, 
$DegComp(c_1) = \frac{\frac{1}{1e^{100}} +  \frac{3}{nVars()} + \frac{8}{nVars()}}{3}$, with $nVars()$ the number of variables of the Boolean formula. 

After giving the necessary definitions (current reference learned clause and Degree of compromise), the following lemma  offers a swifter solution rather than pairwise comparisons, to find relevant clauses based on dominance relationship. 

\begin{lemme}
\label{lemme:minDeg}
Let $c$ be a learned clause having the minimal degree of compromise with respect to the set of learned clauses relevant measures $\cal{M}$, then $c$ is an undominated clause.
\end{lemme}

\begin{preuve}
\label{proof:mindeg}
Let $c$ be a learned clause having the minimal degree of compromise with respect to the set of learned clauses relevant measures $\cal{M}$, we suppose that there exists a learned clause $c'$ that strictly dominates $c$, which means that $\forall m \in \cal{M}$,  $m(c') \succeq m(c)$ and $\exists m' \in \cal{M}$, $m'(c') \succ m'(c)$. Hence, we have $DegComp(c') < DegComp(c)$. The latter inequality contradicts our hypothesis, since $c$ has the minimal degree of compromise with respect to $\cal{M}$.
\end{preuve}

\begin{propriete}
\label{prop:dominance}
Let $\cal{M}$ the set of learned clauses relevant measures, $\forall c, c', c"$ three learned clauses, if $c \succ c'$ and $c' \succ c"$ then $c \succ c"$.
\end{propriete}

During the search process, at each cleaning step of the learned clauses database,  we first find the learned clause $cMin$ having  the minimal degree of compromise with respect to $\cal{M}$. Then, we delete from the learned clauses database all the clauses dominated by $cMin$. 

Searching for all undominated clauses during each cleaning step can be time consuming, such that we only compute the undominated clauses with respect to the reference learned clause during each reduction step.


\subsection{Algorithm}
\label{sec:algo}
In this section, after presenting the general scheme of a deletion strategy of learned clauses ($reduceDB(\Delta)$) adopted by most of the reported solvers,  we propose an algorithm allowing to discover relevant learned clauses by using dominance relationship.

Algorithm \ref{alg:reduceDB} depicts the general scheme of a learned clause deletion strategy ($reduceDB(\Delta)$).
This algorithm first sorts the set of learned clauses according to the defined criterion and then deletes half of the learned clauses. In fact, this algorithm take a learned clauses database of size $n$ and outputs a learned clauses database of size $n/2$.
This is different from our approach which first searches the learned clause having the smallest degree of compromise (called reference learned clause) and then removes all the learned clauses that it dominates.
The algorithm \ref{alg:reduceDB_dom} depicts our learned clause deletion strategy.
It is important to note that the clauses
whose size (number of literals) and LBD are less than or equal to $2$ are not concerned by the dominance relationship. These learned clauses are considered as more relevant and are maintained in the learned clauses database. Hence, the  $minDegComp$ function of our algorithm \ref{alg:reduceDB_dom} looks the learned clause of minimal degree of compromise among the learned clauses of size and LBD greater than $3$.

\restylealgo{ruled}\linesnumbered
\begin{algorithm}[H]
  \caption{\label{alg:reduceDB}Deletion Strategy: {\tt reduceDB} function}
  \SetVline
  \SetInd{0.3em}{0.7em}
  \KwIn{$\Delta$: The learned clauses database of size $n$}
  \KwOut{$\Delta$ The new learned clauses database of size $n/2$}
  \SetKwFunction{TIMEREDUCE}{timeToReduce}
  \SetKwFunction{REDUCE}{reductionOfLearntClauses}
  \SetKwFunction{TRIER}{sortLearntClauses} 
  \SetKwFunction{REMOVECLAUSE}{removeClause}
  \SetKwFunction{SAVECLAUSE}{saveClause}
    
  \SetKwBlock{Begin}{Debut}{Fin}  
  {
      {\TRIER{}} \tcc*{by the defined criterion}
      $limit = n/2$\;
      $ind = 0$\;
      \While{$ind < limit$}
      {
        $clause = \Delta[ind]$ \;
        \If{ $clause.size()> 2$ and $clause.lbd()> 2$} {{\REMOVECLAUSE{}} \;}
        \Else{{\SAVECLAUSE{}} \;}
        $ind++$\;
      }
    \Return $\Delta$ \;
  }
\end{algorithm}

\restylealgo{ruled}\linesnumbered
\begin{algorithm}[H]
  \caption{\label{alg:reduceDB_dom}{\tt reduceDB-Dominance-Relationship} }
  \SetVline
  \SetInd{0.3em}{0.7em}
  \KwIn{$\Delta$: The learned clauses database; $\cal{M}$: a set of relevant measures}
  \KwOut{$\Delta$ The new learned clauses database}
  \SetKwFunction{FUNCTION}{Function}
  \SetKwFunction{REDUCE}{reductionOfLearntClauses}
  \SetKwFunction{MINDEG}{minDegComp}
  \SetKwFunction{DOMINE}{dominates} 
  \SetKwFunction{TRIER}{sortLearntClauses}
  \SetKwFunction{REMOVECLAUSE}{removeClause}
  \SetKwFunction{SAVECLAUSE}{saveClause}
    
  \SetKwBlock{Begin}{Debut}{Fin}  
  {
      {cMin = \MINDEG{$\cal{M}$}} \tcc*{cMin the clause having minimal degree of compromise according to $\cal{M}$}
      $ind = 0$\;
      \While{$ind < |\Delta|$}
      {
        $c = \Delta[ind]$  \tcc*{a learned clause}
        \If{$c.size()> 2$ and $c.lbd()> 2$ and \DOMINE{cMin, c, $\cal{M}$}} {{\REMOVECLAUSE{}} \;}
        \Else{{\SAVECLAUSE{}} \;}
        $ind++$\;
      }
    \Return $\Delta$ \;
  }
  \vspace{0.3cm}
  Function \DOMINE{cMin: a clause, c: a clause, $\cal{M}$} \\
  $i = 0$\;
  \While{$i < |\cal{M}|$}
   { $m = \cal{M}$[i] \tcc*{a relevant measure}
   	 \If{$m(c) \succeq m(cMin)$ } {\Return $FALSE$ \;}
     $i++$\;
   	}
   \Return $TRUE$ \;
  
\end{algorithm}

\section{Experiments}
\label{sec:exps}
For our experiments, we use three relevant measures for the dominance relationship to assess the efficiency of our approach. Notice that the user can choose to combine different other measures.
We use SIZE \cite{Goldberg20071549}, LBD \cite{AudemardS09} and $CVSIDS$ \cite{MiniSat03} measures. All these measures have been proved effective in the literature \cite{MiniSat03,AudemardS09,JabbourLSS14}. It is possible to use more relevant measures, but it should be noted that by adding a measure to $\cal{M}$, the number of relevant learned clauses maintained  may decrease or increase. The decrease can be explained by the fact that a learned clause can be dominated with respect to a set of measures $\cal{M}$ and undominated with respect to $\cal{M'}$ , such that $\cal{M}$  $\subset$ $\cal{M'}$. 
For example, if two learned clauses $c$ and $c'$ are undominated with respect to $\cal{M}$, there is a possibility that one of them dominates the other by removing one measure. The increase can be explained by the fact that a learned clause can be dominated with respect to $\cal{M}$ and undominated with respect to $\cal{M'}$.
For example, consider a learned clause $c$ which dominates another learned clause $c'$
with respect to $\cal{M}$, by adding a measure $m$ to $\cal{M}$, such
that $m(c') \succ m(c)$, then $c'$ is no longer dominated by $c$.

We run the SAT solvers on the 300 instances taken from the last SAT-RACE 2015 and on the 300 instances taken from the last SAT competition 2016.  All the instances are preprocessed by SatElite \cite{satelite} before running the SAT solver.
The experiments are made using Intel Xeon quad-core machines with 32GB of RAM running at 2.66 Ghz. For each instance, we used a timeout of 1 hour of CPU time for the SAT-RACE, and 10000s for the SAT Competition. We integrate our approach in $Glucose$ and made a comparison between the original solver and the one enhanced with the new deletion learned clause strategy using dominance relationship called {\tt DegComp-Glucose}. 

\subsection{Number of solved instances and CPU time}

\begin{figure}[htbp]
\centering
\includegraphics[width=8.5cm]{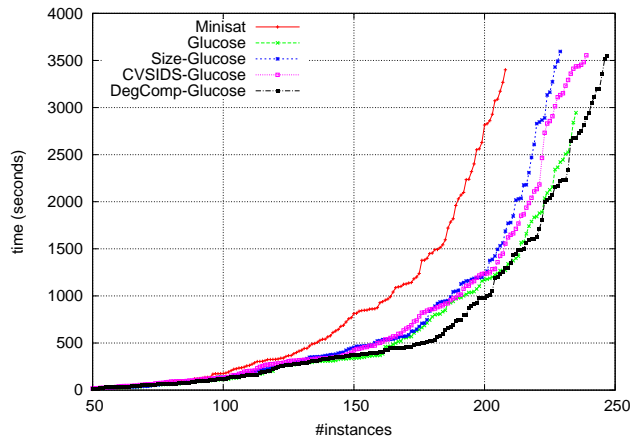}
  \caption{\bf Evaluation on SAT-RACE-2015}
  \label{fig:catusPlot1}
\end{figure}

Table \ref{tab:Compar3} presents results on SAT-RACE. We use the source code of $Glucose$ $3.0$ with the measure $LBD$ (written $LBD$-$Glucose$ or $Glucose$ in what follows). We then replace $LBD$ by each of the other measures : $SIZE$-$Glucose$ that considers the shortest clauses as the most relevant, $CVSIDS$-$Glucose$ that maintains the learned clauses most involved in recent conflict analysis, and finally our proposal $DegComp$-$Glucose$. Table \ref{tab:Compar3} shows the comparative experimental evaluation of the four measures as well as $Minisat$ $2.2$.  In the second column of Table \ref{tab:Compar3}, we give the total number of solved instances (\#Solved). We also mention, the number of instances proven satisfiable (\#SAT) and unsatisfiable (\#UNSAT) in parenthesis. The third column shows the average CPU time in seconds (total time on solved instances divided by the number of solved instances). On the SAT-RACE 2015, our approach $DegComp$-$Glucose$ is more efficient than the others in terms of the number of solved instances (see also Figure \ref{fig:catusPlot1}).
In fact the original solver $Glucose$ solves $236$ instances while it is enhanced with our dominance approach as $12$ more instances are solved. In fact, solving such additional number of instances is clearly significant in practical SAT solving.
The $CVSIDS$-$Glucose$ solver solves $4$ more instances than $Glucose$ $3.0$. $Minisat$ $2.2$ is the worst solver among the five solvers.

\begin{table}[h]
\centering
\begin{tabular}{|l|c|c|}
 \hline
{$Solvers$}  & {\#Solved (\#SAT - \#UNSAT)} & {Average Time}  \\ \hline
{$Minisat$ $2.2$} &{209 (134 - 75)} & {585.19 s}    \\  \hline
{$SIZE$-$Glucose$ }  & {230 ({131} - 99)} & {533.86 s} \\  \hline
{$CVSIDS$-$Glucose$ }  & {240 ({140} - 100)} & {622.23 s} \\  \hline
{$LBD$-$Glucose$}& {236({136} - 100)} & {\bf 481.66 s} \\  \hline
\hline
{$DegComp$-$Glucose$}  & {\bf 248 ({\bf 146} - 102)} & {571.31 s} \\  \hline

\end{tabular}
\caption{Comparative evaluation on {SAT-RACE-2015}.}
\label{tab:Compar3}
\end{table}

Table \ref{tab:Compar4} shows 5 instances of the SAT-RACE 2015 solved by our approach but not solved by $LBD$-$Glucose$, $SIZE$-$Glucose$, nor $CVSIDS$-$Glucose$. The time used to solve those instances may also explain the increase of the average running time of $DegComp$-$Glucose$. In addition we also find that there is none instance solved by all the other solvers and not solved by our approach (as detailed later). This shows on the one hand that the application of dominance between different relevant measures
does not degrade the performance of all the solvers but instead takes advantage of the performance of each relevant measure, considering the SAT-RACE dataset.

\begin{table}[h]
\centering
\begin{tabular}{|l|c|c|c|c|}
 \hline
{$Instances$}  & {LBD} & {SIZE} &  CVSIDS & DegComp\\ \hline
{jgiraldezlevy.2200.9086.08.40.8} &{-} & - & - & {\bf 93.71 s} \\  \hline
{manthey\_DimacsSorterHalf\_37\_3}  & {-} & {-} & - & {\bf 2642.88 s} \\  \hline
{14packages-2008seed.040}  & {-} & {-} & - & {\bf 1713.46 s} \\  \hline
{manthey\_DimacsSorter\_37\_3}& {-} & {-} & {-} & {\bf 2673.39 s} \\  \hline
{jgiraldezlevy.2200.9086.08.40.2}& {-} & {-} & {-} &  {\bf 3195.03 s} \\  \hline
\end{tabular}
\vspace{0.2cm}
\caption{Instances solved by $DegComp$-$Glucose$ and not solved by the others on SAT-RACE.}
\label{tab:Compar4}
\end{table}

Figure \ref{fig:catusPlot1} shows the cumulated time results i.e. the number of instances (x-axis) solved under a given amount of time in seconds (y-axis).
This figure gives for each technique the number of solved instances ($\#instances$) in less than $t$ seconds. It confirms the efficiency of our dominance relationship approach. From  this figure, we can  observe that  $DegComp$-$Glucose$ is generally faster than all the other solvers, even if the average running time of $LBD$-$Glucose$ is the lowest one (see Table \ref{tab:Compar3}). Although $DegComp$-$Glucose$ needs additionnal time to compute the dominance relationship, the quality of the remained clauses on SAT-RACE helps to improve the time needed to solved the instances.

\begin{table}[h]
\centering
\begin{tabular}{|l|c|c|}
 \hline
{$Solvers$}  & {\#Solved (\#SAT - \#UNSAT)} & {Average Time}  \\ \hline
{$Minisat$ $2.2$} &{138 (65 - 73)} & {1194.85 s}    \\  \hline
{$SIZE$-$Glucose$ }  & {156 ({67} - 89)} & {1396.73 s} \\  \hline
{$CVSIDS$-$Glucose$ }  & {165 ({67} - \bf 98)} & { 1368.99 s} \\  \hline
{$LBD$-$Glucose$ }& {165 ({68} - { 97})} & {\bf 1142.33 s} \\  \hline
\hline
{$DegComp$-$Glucose$}  & {164 ({\bf 69} - 95)} & {1456.34 s} \\  \hline

\end{tabular}
\vspace{0.2cm}
\caption{Comparative evaluation on {SAT-Competition-2016}.}
\label{tab:Compar8}
\end{table}

\begin{figure}[htbp]
\centering
\includegraphics[width=8.5cm]{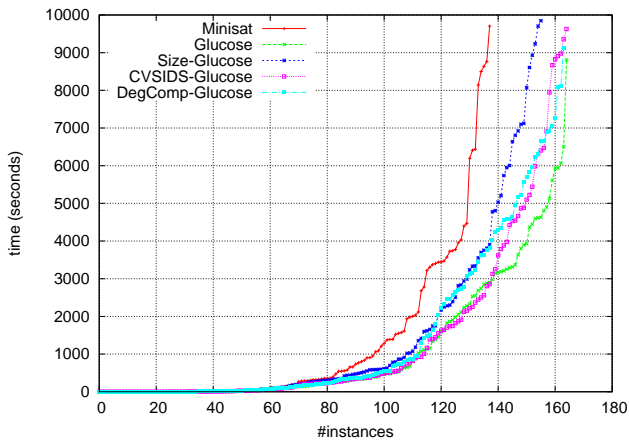}
  \caption{\bf Evaluation on SAT competition 2016}
  \label{fig:catusPlot2}
\end{figure}

Table \ref{tab:Compar8} presents results on the instances of the SAT Competition 2016. Here $LBD$-$Glucose$ and $CVSIDS$-$Glucose$  solve one more instance than $DegComp$-$Glucose$ which remains competitive, and solves the greatest number of satisfiable instances. Figure \ref{fig:catusPlot2} presents the cumulated time results on the instances of the SAT competition 2016. It comes out from this second dataset that $LBD$-$Glucose$ is more efficient than the others including our approach which remains competitive wrt the number of solved instances. 

This outcome gives credit to the NO FREE Lunch theorem \cite{DolpertM97}. We also think that the aggregated function may not be unique for all the datasets, such that it is necessary to explore the efficient combination of the prefered measures.


\subsection{Common solved instances}

In table \ref{tab:Compar5}, the intersection between two relevant measures gives the number of common instances solved by each measure. For example, $LBD$ and $SIZE$ solved 219 instances in common, while $234$ instances are solved by $LBD$ and $DegComp$.
We can see than our approach solves the largest number of instances in common with each of the aggregated measures. More precisely, the number of common instances solved with another measure is lower than the number of common instances solved with our approach.  

\begin{table}[h]
\centering
\begin{tabular}{|l||c|c|c|c|}
 \hline
{$Measures$}  & {LBD}       & {SIZE}     &  CVSIDS &   DegComp\\ \hline \hline
{LBD}        & {\bf 236}   &         &       &    {\bf 234} \\  \hline
{SIZE}       &   219       & {\bf 230}  &      &    {\bf 225} \\  \hline
{CVSIDS}     &   233       &221        &  {\bf 240}   &    {\bf 238} \\  \hline
{DegComp}    &         &    &      &    {\bf 248} \\  \hline
\end{tabular}
\vspace{0.2cm}
\caption{Common solved Instances from SAT-RACE-2015.}
\label{tab:Compar5}
\end{table}


To get more details, Table \ref{tab:Compar5bis} gives the number of instances commonly solved by the considered relevant measures. This table allows to see the number of common instances solved by one, two, three or four measures. For example, there are $218$ common instances solved by the four deletion strategies, while $44$ instances are not solved by none of them. We can observe that $1$, $1$, $5$, and $5$ are the number of instances solved alone by respectively $LBD$ and $CVSIDS$, $SIZE$ and $DegComp$. 
Moreover, there is no instance solved by the three strategies ($LBD$, $SIZE$ and $CVSIDS$) and not solved by our approach $DegComp$.

\begin{table}[h]
\centering
\begin{tabular}{|l|c|c|c|c|c|}
 \hline
{$Measures$}  &  & \multicolumn{2}{c|}{DegComp} &  \multicolumn{2}{c|}{$\neg$DegComp}  \\ \hline
    &    & {CVSIDS}  &  {$\neg$CVSIDS} & {CVSIDS}  &   {$\neg$CVSIDS}  \\ \hline
  {LBD} &  {SIZE}    & 218  &  1 & {\bf 0 } & 0 \\  \cline{2-6}
   & {$\neg$SIZE}    & 1     &  1  & 1  &  1 \\  \hline
 {$\neg$ LBD} & {SIZE}   & 3  & 3  & 0  &  { \bf 5 }  \\  \cline{2-6}
  &  {$\neg$SIZE}    &  3  &  {\bf 5}  & 1 &  44 \\  \hline

\end{tabular}
\vspace{0.2cm}
\caption{Detailed of common instances with SAT-RACE.}
\label{tab:Compar5bis}
\end{table}

\subsection{Combined measures}

Table \ref{tab:Compar6} gives the number of instances solved with our dominance approach wrt the measures used in the dominance relations. From this table, we can see that the number of instances solved by using two measures (instead of three) in the dominance relationship is always lower than the number of instances solved  ($248$) by using three measures.

\begin{table}[h!]
\centering
\begin{tabular}{|l||c|c|c|c|}
 \hline
{$Measures$}  & {LBD}       & {SIZE}     &  CVSIDS &   DegComp\\ \hline \hline
{LBD}        & {\bf 236}   &      &        & \cellcolor{gray}  \\ \cline{1-4}
{SIZE}       &    223      & {\bf 230}  &      & \cellcolor{gray}   \\  \cline{1-4}
{CVSIDS}     &     239     &    242     &  {\bf 240}   & \cellcolor{gray}     \\  \hline
{DegComp}    &  \multicolumn{3}{l|}{\cellcolor{gray}}  & {\bf 248} \\  \hline
\end{tabular}
\vspace{0.2cm}
\caption{Combining two measures on SAT-RACE-2015.}
\label{tab:Compar6}
\end{table}




\subsection{Percentage of deleted clauses}

During our experiments, we compute at each reduction step of instance resolution, the percentage of deleted clauses i.e the number of dominated clauses (which are the removed) over the total number of learned clauses during this step. This allows to obtain an average percentage of deleted learned clauses per solved instance. By taking all the solved instances of the SAT-RACE 2015, the average of the average percentage of deleted learned clauses is equal to $0.36$ with a standard deviation of $0.16$.

\begin{figure}[h!]
\centering
\includegraphics[width=6cm,angle=270]{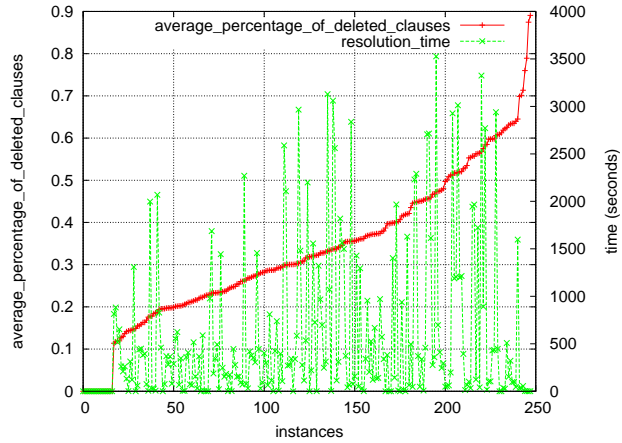}
  \caption{\bf Average percentage of deleted clauses and resolution time for each instance.}
  \label{fig:catusPlot-deletedClauses}
\end{figure}

\begin{figure}[h!]
\centering
\includegraphics[width=6cm,angle=270]{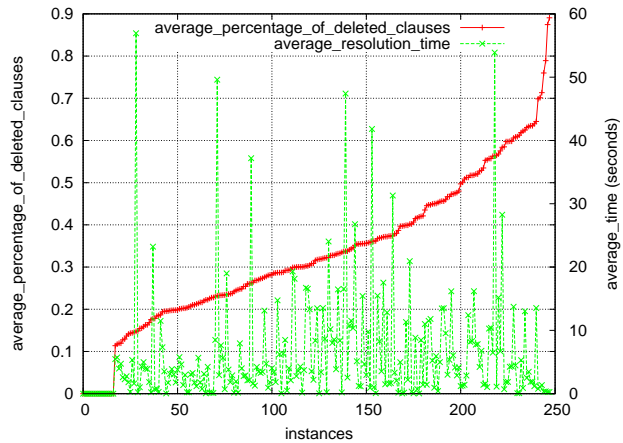}
  \caption{\bf Average percentage of deleted clauses and average resolution time for each instance.}
  \label{fig:catusPlot-averageDeletedClause}
\end{figure}

Figures \ref{fig:catusPlot-deletedClauses} and \ref{fig:catusPlot-averageDeletedClause} plot for each solved instance of the SAT-RACE 2015 (X-axis), the average percentage of deleted learned clauses (red curve with left-Y-axis) against respectively the total resolution time and the average resolution time (green curve with right-Y-axis). For each solved instance, the average resolution time is obtained by dividing its total resolution time by the number of reductions made before solving the instance.

The (red) curve of the average percentages of deleted learned clauses exhibits a high variation of the percentages of reduction from $0.11$ to $0.89$, with an average value equals to $0.36$ with a standard deviation of $0.16$. It comes out from this figure that the average percentage of deleted learned clauses is less than $50\%$ on $200$ instances among $248$ solved instances. Our current strategy which uses only one undominated clause at each step is satisfactory wrt the running time, even if it can be possible to extend this stategy to a reduction with many undominated clauses.
The curve of the average percentages of deleted learned clauses also shows $17$ instances having the average percentage of deleted learned clauses equal to $0$. These $17$ instances correspond to the instances solved by the solver without having to reduce the learned clauses database.

Figure \ref{fig:catusPlot-deletedClauses} shows on the one hand, the instances whose resolution times are small but with a high average percentage of deleted learned clauses, and on the other hand, the instances whose resolution times are high but with a low average percentage of deleted learned clauses. The same remark is also valid with figure \ref{fig:catusPlot-averageDeletedClause} where we use the average resolution time instead of the total resolution time.

This clearly shows that the number of deleted learned clauses at each reduction step is not the only component that impacts the resolution time. Other key components of modern CDCL SAT solver such as the restart policies \cite{Gomes1998}  and the activity-based variable selection heuristics \cite{Moskewicz01} also have an influence on the resolution time.

It should be noted that if the solver \textit{Glucose} keeps all  the learned clauses (no learned clauses deleted) throughout the resolution process, it solves only $203$ instances on the SAT-RACE 2015 instances in $1$ hour ($33$ instances less than the original solver and $45$ instances of less than the solver integrating our dominance approach).
On the instances of the 2016 SAT competition, keeping all the learned clauses  during the resolution process, the solver glucose solves only $131$ in $10000$ seconds ($34$ instances less than the original solver and $33$ instances of less than the solver integrating our approach).

This confirms the need to eliminate certain learned clauses (those deemed irrelevant) during the resolution process, and otherwise the interest of the learned clauses removal problem in CDCL SAT solvers.

\section{Conclusion and Future Works}
\label{sec:conclusion}

In this paper, we propose an approach that addresses the learned clauses database management problem.
We have shown that the idea of dominance relationship between relevant measures is a nice way to take profit of each measure.
This approach is not hindered by the abundance of relevant measures which has been the issue of several works. 
The proposed approach avoids another non-trivial problem which is the amount of learned clauses to be deleted at each reduction step of the learned clauses database.
The experimental results show that exploiting the dominance relationship improves the performance of CDCL SAT solver, at least on the SAT-RACE 2015. For the case of SAT-Competition, we still have to find a good dominance relation. The instances categories might also be an issue which should be explored.

To the best of our knowledge, this is the first time that dominance relationship has been used in the satisfiability domain to improve the performance of a CDCL SAT solver. 
Our approach opens interesting perspectives. In fact, any new relevant measure of learned clauses can be integrated into the dominance relationship.

\section*{Acknowledgements}
The authors would like to thank Auvergne-Rh\^one-Alpes region and European Union for their financial support through the European Regional Development Fund (ERDF). 
The authors would also like to thank CRIL (Lens Computer Science Research Lab) for providing them computing server and the authors of $Glucose$ solver for making available the source code of  their solver.

\bibliographystyle{plain}
\bibliography{biblio,satBib}

\begin{thebibliography}{10}

\bibitem{AnsoteguiGLS15}
Carlos Ans{\'{o}}tegui, Jes{\'{u}}s Gir{\'{a}}ldez{-}Cru, Jordi Levy, and
  Laurent Simon.
\newblock Using community structure to detect relevant learnt clauses.
\newblock In {\em {SAT} 2015}, pages 238--254, 2015.

\bibitem{AudemardS09}
G.~Audemard and L.~Simon.
\newblock Predicting learnt clauses quality in modern sat solvers.
\newblock In {\em IJCAI'09}, pages 399--404, 2009.

\bibitem{AudemardLMS11}
Gilles Audemard, Jean-Marie Lagniez, Bertrand Mazure, and Lakhdar Sais.
\newblock On freezing and reactivating learnt clauses.
\newblock In {\em SAT'2011}, pages 188--200, 2011.

\bibitem{Bayardo96acomplexity}
Roberto~J. Bayardo and Daniel~P. Miranker.
\newblock A complexity analysis of space-bounded learning algorithms for the
  constraint satisfaction problem.
\newblock In {\em In Proceedings of the Thirteenth National Conference on
  Artificial Intelligence}, pages 298--304, 1996.

\bibitem{Bayardo97}
Roberto~J. {Bayardo, Jr.} and Robert~C. Schrag.
\newblock Using {CSP} look-back techniques to solve real-world {SAT} instances.
\newblock In {\em AAAI}, pages 203--208, 1997.

\bibitem{Lingeling12}
A.~Biere.
\newblock Lingeling and friends entering the sat challenge 2012.
\newblock In A.~Balint, A.~Belov, A.~Diepold, S.~Gerber, M.~Jarvisalo, , and
  C.~Sinz (editors), editors, {\em Proceedings of SAT Challenge 2012: Solver
  and Benchmark Descriptions}, pages 33--34, University of Helsinki, 2012. vol.
  B-2012-2 of Department of Computer Science Series of Publications B.

\bibitem{BorzsonyiKS01}
Stephan B{\"{o}}rzs{\"{o}}nyi, Donald Kossmann, and Konrad Stocker.
\newblock The skyline operator.
\newblock In {\em Proceedings of the 17th International Conference on Data
  Engineering, April 2-6, 2001, Heidelberg, Germany}, pages 421--430, 2001.

\bibitem{BoukerSYN14}
Slim Bouker, Rabie Saidi, Sadok~Ben Yahia, and Engelbert~Mephu Nguifo.
\newblock Mining undominated association rules through interestingness
  measures.
\newblock {\em International Journal on Artificial Intelligence Tools}, 23(4),
  2014.

\bibitem{Davis62}
Martin Davis, Gearge Logemann, and Donald~W. Loveland.
\newblock A machine program for theorem-proving.
\newblock {\em Communications of the ACM}, 5(7):394--397, 1962.

\bibitem{satelite}
Niklas E{\'e}n and Armin Biere.
\newblock Effective preprocessing in sat through variable and clause
  elimination.
\newblock In {\em SAT'05}, pages 61--75, 2005.

\bibitem{MiniSat03}
Niklas E{\'{e}}n and Niklas S{\"{o}}rensson.
\newblock An extensible sat-solver.
\newblock In {\em {SAT} 2003}, pages 502--518, 2003.

\bibitem{Goldberg20071549}
Eugene Goldberg and Yakov Novikov.
\newblock Berkmin: A fast and robust sat-solver.
\newblock {\em Discrete Applied Mathematics}, 155(12):1549 -- 1561, 2007.

\bibitem{Gomes1998}
Carla~P. Gomes, Bart Selman, and Henry~A. Kautz.
\newblock Boosting combinatorial search through randomization.
\newblock In {\em AAAI/IAAI}, pages 431--437, 1998.

\bibitem{GuoJLS14}
Long Guo, Sa{\"{\i}}d Jabbour, Jerry Lonlac, and Lakhdar Sais.
\newblock Diversification by clauses deletion strategies in portfolio parallel
  {SAT} solving.
\newblock In {\em {ICTAI} 2014}, pages 701--708, 2014.

\bibitem{JabbourLSS14}
Sa{\"{\i}}d Jabbour, Jerry Lonlac, Lakhdar Sais, and Yakoub Salhi.
\newblock Revisiting the learned clauses database reduction strategies.
\newblock {\em CoRR}, abs/1402.1956, 2014.

\bibitem{KatebiSS11}
Hadi Katebi, Karem~A. Sakallah, and Jo{\~{a}}o P.~Marques Silva.
\newblock Empirical study of the anatomy of modern sat solvers.
\newblock In {\em {SAT} 2011}, pages 343--356, 2011.

\bibitem{Moskewicz01}
Matthew~W. Moskewicz, Conor~F. Madigan, Ying Zhao, Lintao Zhang, and Sharad
  Malik.
\newblock Chaff: Engineering an efficient sat solver.
\newblock In {\em 38th Design Automation Conference (DAC'01)}, pages 530--535,
  2001.

\bibitem{PipatsrisawatD09}
Knot Pipatsrisawat and Adnan Darwiche.
\newblock On the power of clause-learning sat solvers with restarts.
\newblock In {\em (CP'09)}, pages 654--668, 2009.

\bibitem{Marques-Silva96}
Jo{\~a}o P.~Marques Silva and Karem~A. Sakallah.
\newblock Grasp - a new search algorithm for satisfiability.
\newblock In {\em International Conference on Computer-Aided Design
  (ICCAD'96)}, pages 220--227, 1996.

\bibitem{Marques-SilvaS99}
Jo{\~{a}}o P.~Marques Silva and Karem~A. Sakallah.
\newblock {GRASP:} {A} search algorithm for propositional satisfiability.
\newblock {\em {IEEE} Trans. Computers}, 48(5):506--521, 1999.

\bibitem{SouletRPC11}
Arnaud Soulet, Chedy Ra{\"{\i}}ssi, Marc Plantevit, and Bruno Cr{\'{e}}milleux.
\newblock Mining dominant patterns in the sky.
\newblock In {\em {ICDM} 2011}, pages 655--664, 2011.

\bibitem{DolpertM97}
David Wolpert and William~G. Macready.
\newblock No free lunch theorems for optimization.
\newblock {\em {IEEE} Trans. Evolutionary Computation}, 1(1):67--82, 1997.

\end{thebibliography}
\end{document}